\documentclass[letterpaper]{article} 
\usepackage{aaai2026}  
\usepackage{times}  
\usepackage{amsmath}
\usepackage{amssymb}
\usepackage{helvet}  
\usepackage{courier}  
\usepackage[hyphens]{url}  
\usepackage{graphicx} 
\usepackage{natbib}  
\usepackage{caption} 
\usepackage{algorithm}
\usepackage{algorithmic}
\usepackage{enumitem}
\usepackage{graphicx}
\usepackage{xcolor}

\usepackage{newfloat}
\usepackage{listings}
\DeclareCaptionStyle{ruled}{labelfont=normalfont,labelsep=colon,strut=off} 
\floatstyle{ruled}
\newfloat{listing}{tb}{lst}{}
\floatname{listing}{Listing}
\title{Large Language Models Reasoning Abilities Under Non-Ideal Conditions After RL-Fine-Tuning}
\author{
    Chang Tian\textsuperscript{\rm 1},
    Matthew B. Blaschko\textsuperscript{\rm 1},\\
    Mingzhe Xing\textsuperscript{\rm 2},
    Xiuxing Li\textsuperscript{\rm 3},
    Yinliang Yue\textsuperscript{\rm 2},
    Marie-Francine Moens\textsuperscript{\rm 1}
}
\affiliations{
    \textsuperscript{\rm 1}KU Leuven\\

    \textsuperscript{\rm 2}Zhongguancun Lab\\
    \textsuperscript{\rm 3}Beijing Institute of Technology\\
    namechangtian@163.com 
}

\usepackage{bibentry}

\begin{document}

\maketitle

\begin{abstract}
Reinforcement learning (RL) has become a key technique for enhancing the reasoning abilities of large language models (LLMs), with policy gradient algorithms dominating the post‑training stage because of their efficiency and effectiveness. However, most existing benchmarks evaluate large language model reasoning under idealized settings, overlooking performance in realistic, non-ideal scenarios. We identify three representative non-ideal scenarios with practical relevance: summary inference, fine‑grained noise suppression, and contextual filtering.
We introduce a new \textbf{research direction} guided by brain science findings that human reasoning remains reliable under imperfect inputs. We formally define and evaluate these challenging scenarios. We fine‑tune three LLMs and a state-of-the-art large vision-language model (LVLM) using RL with a representative policy gradient algorithm and then test their performance on eight public datasets. 
Our results reveal that, while RL fine-tuning improves baseline reasoning under idealized settings, performance declines significantly across all three non-ideal scenarios, exposing critical limitations in advanced reasoning capabilities. Although we propose a scenario-specific remediation method, our results suggest current methods leave these reasoning deficits largely unresolved. This work highlights that the reasoning abilities of large models (LMs) are often overstated and underscores the importance of evaluating models under non-ideal scenarios. All code and data will be released at https://github.com/xxxx.
\end{abstract}
\section{Introduction}
Large language models have exhibited the attractive performance in reasoning tasks~\citep{kumar2025llm, li2021paint4poem,tian2025using,tian2024fighting}.
Leveraging reinforcement learning has become pivotal in refining LLM reasoning~\citep{schulman2017proximal}.
Recently, some research works integrate Monte Carlo Tree Search (MCTS) into RL pipelines to enhance exploratory capacity and uncover superior reasoning trajectories~\citep{xie2024monte}.
Policy gradient approaches, in contrast, streamline learning by directly optimizing policy parameters via reward feedback, bypassing extensive rollout requirements~\citep{guo2025deepseek}. Owing to their scalability and efficiency, policy gradient approaches now dominate post-training strategies for LLMs~\citep{du2025survey}.
A growing body of work demonstrates that RL-fine-tuned LLMs outperform baselines across benchmarks like GSM8K, MATH, and commonsense evaluations~\citep{wang2024reinforcement}.

However, such assessments are typically conducted under ideal, noise-free conditions. As a result, they primarily substantiate improvements in basic reasoning abilities but fall short of validating advanced reasoning. 
In contrast, real-world reasoning requires advanced reasoning abilities to summarize multiple possibilities and filter out noisy information.

Recent advances in brain science have shown that advanced human reasoning involves integrating diverse information, suppressing irrelevant signals, and making contextually grounded decisions.
The advances are published in Nature and Science~\citep{deary2010neuroscience,daume2024control,tune2021neural,paluch2025unattended}. 

Humans apply advanced reasoning abilities daily: detectives weigh alternative possibilities to identify suspects, clinicians discern fine-grained symptom patterns to reach accurate diagnoses, and financial analysts sift extensive information streams to isolate relevant evidence before formulating investment plans. As humans increasingly use large models for assistance~\citep{tian2024generic}, it is necessary to evaluate and enhance large models' advanced reasoning abilities in realistic, non-ideal scenarios so large models can provide support comparable to that of human experts.

\begin{figure}[t]
    \centering
    \includegraphics[width=\columnwidth]{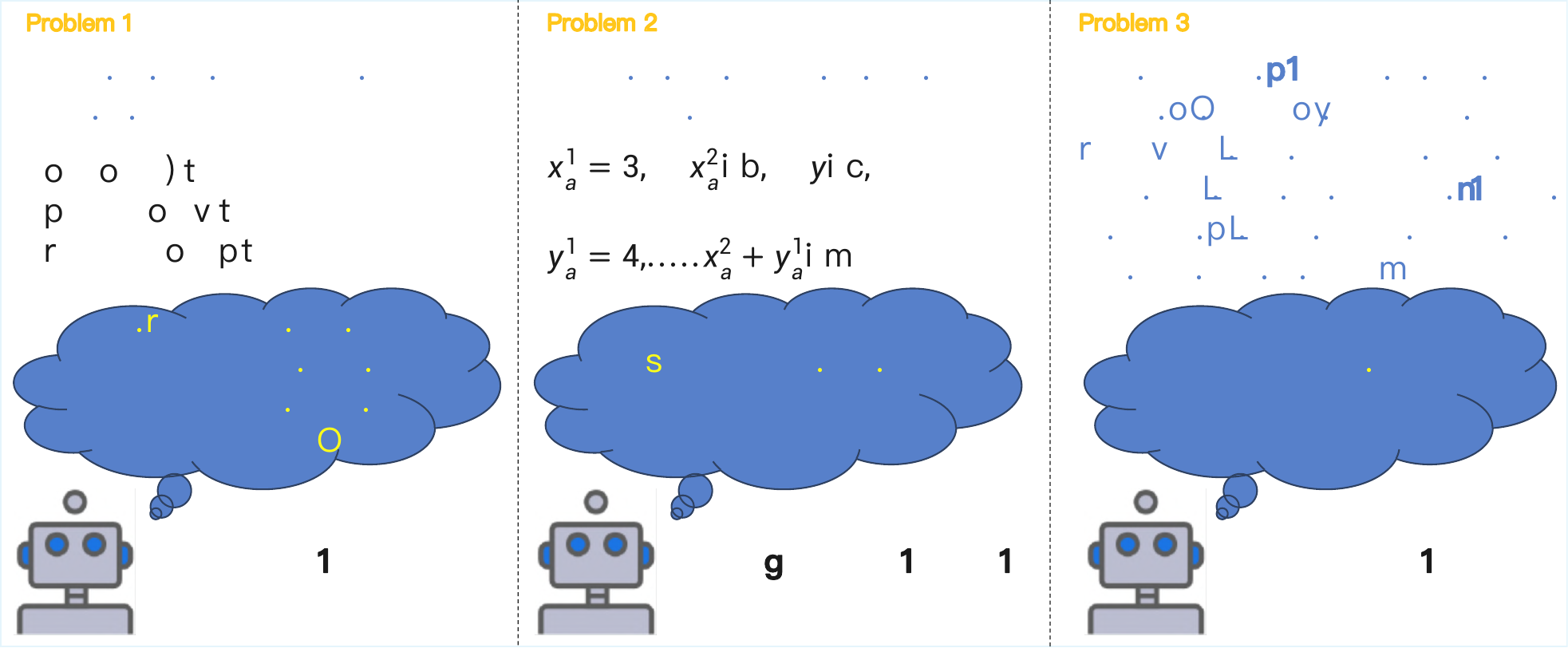}
    \caption{Illustration of problem settings used to evaluate model reasoning performance under non-ideal scenarios.}
    \label{fig:problem_statement}
\end{figure}

Inspired by insights from brain science, we have designed three evaluation tasks to assess the reasoning abilities of LLMs under the non-ideal scenarios illustrated in Figure~\ref{fig:problem_statement}:
\begin{itemize}
    \item \textbf{Summary Inference}. Assume each candidate is correct, analyze accordingly, and aggregate the results into a single conclusion.
    \item \textbf{Fine-grained Noise Suppression}. Detect and remove fine-grained noise.
    \item \textbf{Contextual Filtering}. Discard irrelevant context to maintain reasoning.
\end{itemize}

These tasks are grounded in neuroscientific phenomena and aim to probe advanced reasoning abilities beyond conventional benchmarks.

To systematically investigate these scenarios, we conduct empirical evaluations with three RL–fine-tuned LLMs and one RL–fine-tuned large vision-language model (LVLM). We further explore remediation strategies by introducing variations during training and evaluation. Specifically, we manipulate two factors:
(1) Format reward, which incentivizes the model to consider all possibilities during training;
(2) Example guidance, where we provide a guided example during both training and evaluation to highlight key information and suppress distractions. This results in a spectrum of model variants formed by different training–evaluation stage combinations as shown in Figure~\ref{fig:model_figure}. 

Our contributions are fourfold:
\begin{enumerate}
    \item \textbf{Inspired by brain science}, we propose a new research direction: examining the reasoning abilities of RL-fine-tuned large models under non-ideal scenarios, which has been insufficiently explored in prior work.
    
    \item \textbf{Through systematic evaluation} of four RL-fine-tuned LMs and their variants across eight datasets, we demonstrate that RL-fine-tuned LMs exhibit significant performance degradation under non-ideal scenarios, in stark contrast to their performance in ideal conditions.
    
    \item \textbf{We design effective remediation strategies} tailored to each scenario by manipulating the format reward and the example guidance. These strategies also suggest directions for future model improvements. These interventions further confirm that RL-fine-tuned LMs possess substantial deficits in advanced reasoning, suggesting their reasoning capabilities have been overstated in prior literature.
    
    \item \textbf{We publicly release high-quality, novel evaluation datasets} designed to assess LM performance under noisy conditions, including fine-grained distractors and irrelevant contextual information, providing valuable resources for future research.
\end{enumerate}
\section{Related Work}
\subsection{RL for Enhancing LLM Reasoning}
Reinforcement learning techniques for enhancing LLM reasoning can be categorized into two groups: Monte Carlo--based methods and policy gradient--based methods. Recent works have explored Monte Carlo techniques to improve reasoning quality~\citep{zheng2025monte}.  
\citet{xie2024monte} proposed MCTS-DPO, which integrates Monte Carlo Tree Search (MCTS) generated step-level preference data into Direct Preference Optimization.
Similarly, ReST-MCTS~\citep{zhang2024rest} employs process-level reward estimation via MCTS to guide iterative self-training.  
VinePPO~\citep{kazemnejad2024vineppo} introduces a PPO variant that replaces value networks with unbiased Monte Carlo--based credit assignment.

Policy gradient methods have also achieved strong results in reasoning tasks~\citep{wang2024reinforcement}. PPO~\citep{schulman2017proximal}, GRPO~\citep{guo2025deepseek}, and DAPO~\citep{yu2025dapo} have shown notable effectiveness.  
\citet{xu2025training} proposed EM Policy Gradient, framing reasoning as an expectation--maximization problem over rationale trajectories.  
\citet{zhang2025design} introduced RPG, a KL-regularized policy gradient framework that stabilizes training.

Compared to Monte Carlo methods, policy gradient approaches avoid expensive rollouts, offering superior computational efficiency~\citep{shao2024deepseekmath}. In this work, we adopt a policy gradient method, GRPO, to fine-tune LLMs for advanced reasoning.
\subsection{Assessing the Performance of LLMs}
The evaluation of LLM performance has progressed through robust benchmarks and contamination-sensitive evaluations. The Massive Multitask Language Understanding benchmark (MMLU)~\citep{hendrycks2020measuring} remains a standard for evaluating reasoning across 57 subjects. However, concerns about data leakage have grown, with models achieving inflated scores. \citet{deng2024investigating} investigate this contamination and introduce detection methods such as retrieval-based overlap analysis and Testset Slot Guessing. Statistical approaches like ConStat~\citep{dekoninck2024constat} quantify contamination risks via performance differentials across benchmark variants. To address prompt variability and reasoning depth, MMLU-Pro~\citep{wang2024mmlu} introduces harder questions and improved robustness.

Differently, our work explores a new research direction: evaluating LLM reasoning under non-ideal scenarios.
\section{Background}
To analyze the reasoning performance of LLMs under non-ideal scenarios after fine-tuning with a policy gradient algorithm, we first introduce the foundational policy gradient method. We then present Group Relative Policy Optimization (GRPO)~\citep{guo2025deepseek}, a recent and widely adopted policy gradient variant known for its effectiveness and GPU memory efficiency~\citep{shao2024deepseekmath}. In this work, our LLMs are fine-tuned using GRPO.
\subsection{Policy Gradient Algorithm}
The policy gradient algorithm optimizes a \textit{parameterized policy} \( \pi_\theta \) directly, rather than learning a value function and extracting a policy from it~\citep{sutton1998introduction, tian2022anti}.

For an episodic Markov Decision Process~\citep{puterman1990markov}, the return from time step \( t \) is defined as \(G_t = \sum_{k \ge t} \gamma^{k - t} r_k\),
and the optimisation target is
\begin{equation}
J(\theta) = \mathbb{E}_{\tau \sim \pi_\theta} \left[ G_0 \right],
\end{equation}
which represents the expected discounted return under the current policy \( \pi_\theta \), where \( \tau \) denotes a full episode sampled from \( \pi_\theta \).

The policy-gradient theorem gives:
\begin{equation}
\nabla_\theta J(\theta) = \mathbb{E}_{\tau \sim \pi_\theta} \left[ \sum_{t=0}^{T-1} \nabla_\theta \log \pi_\theta(a_t \mid s_t) A_t \right],
\end{equation}
where \(A_t = Q^{\pi}(s_t, a_t) - b(s_t)\)
is the \textit{advantage}, \( b(s_t) \) is any baseline function independent of the action \( a_t \), and \( s_t \) is the environment state at time step $t$.

\textbf{Loss formulation}.
In practice, one minimises:
\begin{equation}
\mathcal{L}_{\mathrm{PG}}(\theta) = - \mathbb{E}_{\tau \sim \pi_\theta} \left[ \sum_{t=0}^{T-1} \log \pi_\theta(a_t \mid s_t) A_t \right],
\end{equation}
so that stochastic gradient descent on \( \mathcal{L}_{\mathrm{PG}} \) performs gradient ascent on \( J(\theta) \).
\subsection{Group Relative Policy Optimization}
GRPO is proposed as a representative variant of policy gradient methods~\citep{shao2024deepseekmath, guo2025deepseek}, which eliminates the need for a value model, which is typically as large as the policy model, and instead estimates the baseline using group scores~\citep{vojnovic2025alignment}. 

Specifically, for each question \( q \), GRPO samples a group of outputs \( \{o_1, o_2, \cdots, o_G\} \) from the old policy \( \pi_{\theta_\text{old}} \) and then optimizes the policy model \( \pi_\theta \) by maximizing the following objective:
{\small
\begin{multline}
\mathcal{J}_{\text{GRPO}}(\theta) = \mathbb{E}_{q \sim P(Q), \{o_i\}_{i=1}^{G} \sim \pi_{\theta_\text{old}}(O|q)} \biggl[ \\
\frac{1}{G} \sum_{i=1}^{G} \biggl( \min\left( \frac{\pi_\theta(o_i|q)}{\pi_{\theta_\text{old}}(o_i|q)} A_i, \right. \\
\left. \text{clip}\left( \frac{\pi_\theta(o_i|q)}{\pi_{\theta_\text{old}}(o_i|q)}, 1 - \epsilon, 1 + \epsilon \right) A_i \right) \\
- \beta \, \mathrm{D}_{\mathrm{KL}}(\pi_\theta \| \pi_{\text{ref}}) \biggr) \biggr].
\label{eq:grpo_objective}
\end{multline}
}
\begin{equation}
\mathrm{D}_{\mathrm{KL}}(\pi_\theta \| \pi_{\text{ref}}) = 
\frac{\pi_{\text{ref}}(o_i|q)}{\pi_\theta(o_i|q)} - 
\log \frac{\pi_{\text{ref}}(o_i|q)}{\pi_\theta(o_i|q)} - 1,
\label{eq:grpo_kl}
\end{equation}

where \( \epsilon \) and \( \beta \) are hyperparameters, \( q \) is a question, and \( P(Q) \) denotes the probability distribution over the set of possible questions.
The term \( A_i \) represents the advantage, which is computed using a group of rewards \( \{r_1, r_2, \ldots, r_G\} \) corresponding to the outputs within each group:

\begin{equation}
A_i = \frac{r_i - \text{mean}(\{r_1, r_2, \cdots, r_G\})}{\text{std}(\{r_1, r_2, \cdots, r_G\})}.
\label{eq:grpo_advantage}
\end{equation}


    
    

\section{Method}
\begin{figure*}[t!]
    \centering
    \includegraphics[width=0.985\textwidth]{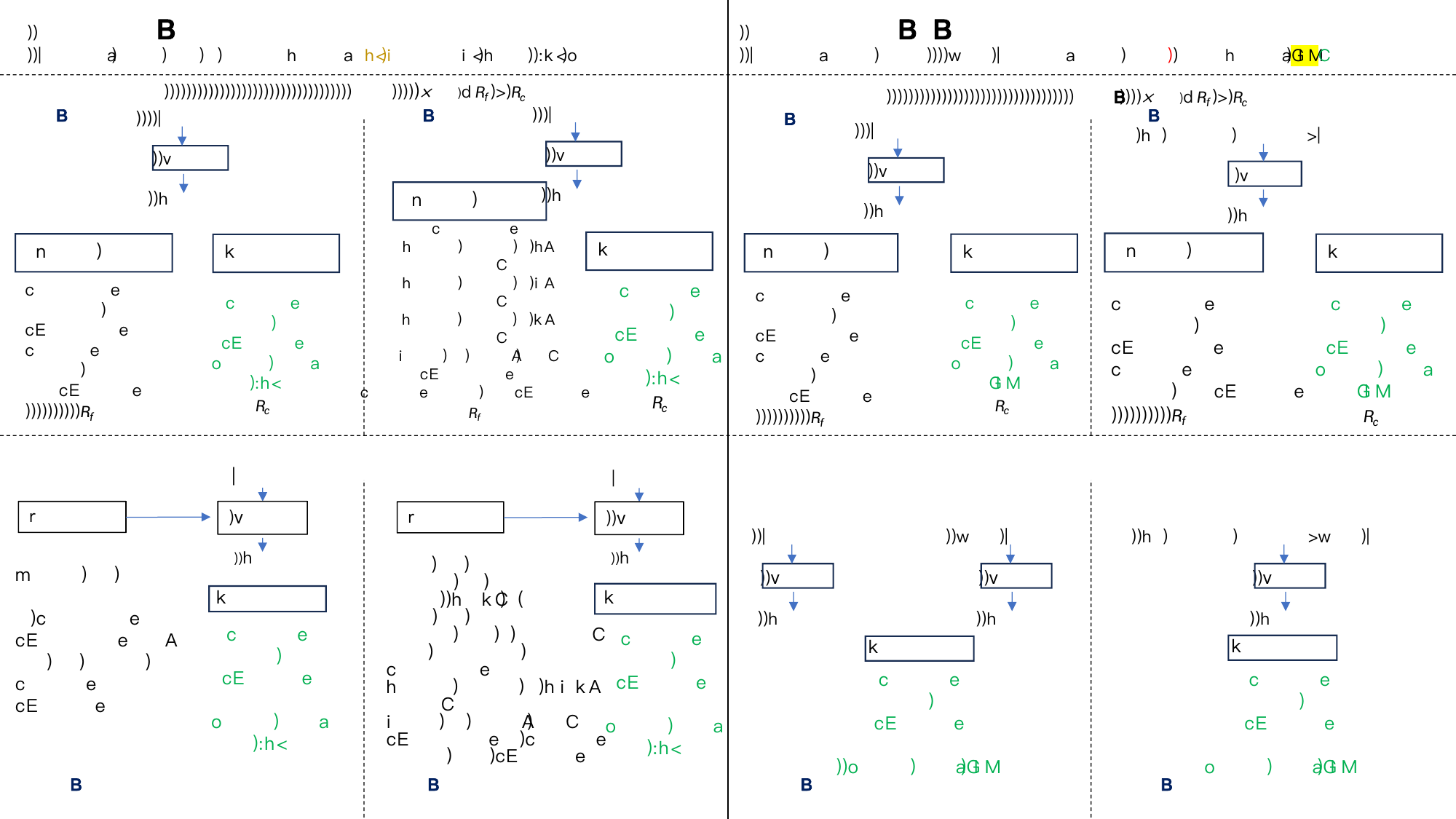}
    \caption{Overview of the training and evaluation stages used to construct the RL-fine-tuned model and its variants. The reward \( R_{\text{train}} \) corresponds to a specific reward \( r_i \) selected from a group of \( G \) rewards, as defined in Equation~\ref{eq:grpo_advantage}. In the training process, \( R_{\text{train}} \) is defined as the sum of the output format reward \( R_f \) and the output correctness reward \( R_c \).}
    \label{fig:model_figure}
\end{figure*}
In this section, we categorize the RL-fine-tuned models and their variants into two groups, as illustrated in Figure~\ref{fig:model_figure}. One group is designed to summarize inference, while the other focuses on reasoning in noisy scenarios, which include both fine-grained noise and irrelevant contextual information.
\subsection{Summary Inference}
As shown in the left part of Figure~\ref{fig:model_figure}, the training process is divided into Stage A and Stage C based on different \textbf{output format rewards}, while the evaluation process is divided into Stage B and Stage D according to the \textbf{instruction requirements} applied during evaluation.

Concretely, in Stage A, we train the model to place the reasoning process within \texttt{<reasoning>}...\texttt{</reasoning>} tags, and to place the predicted answer within \texttt{<answer>}...\texttt{</answer>} tags. In stage C, we train the model to assume each possible answer as correct in turn, perform a brief analysis for each, and then perform summarizing reasoning based on all intermediate analyses. The entire reasoning process is enclosed within the \texttt{<reasoning>}...\texttt{</reasoning>} tags. All format checks in Stage A and Stage C are enforced using regular expressions.

Similarly, in the evaluation process, the instructions for Stage D explicitly require the model to perform a brief analysis for each option followed by the summarized inference. In contrast, Stage B does not impose this requirement.

Based on different combinations of training and evaluation stages, we present the RL-fine-tuned model and its key variants as follows:

\textbf{Model-StageAB}: This is the standard RL-fine-tuned model commonly used in the current research community. Its effectiveness is primarily evaluated based on performance in ideal test settings (Stage B), which has led to optimistic conclusions about the benefits of RL fine-tuning.

\textbf{Model-StageAD}: This variant applies an explicit instruction during evaluation (Stage D), requiring the model to consider each option and summarize the inference. It is designed to assess the model’s reasoning ability under non-ideal conditions. 
When more information is considered, a model with advanced reasoning capabilities is expected to maintain or improve its performance. Otherwise, performance degradation may occur due to increased complexity and confusion. 
The math explanation is provided in Appendix Section Summary Inference.

\textbf{Model-StageCD}: This variant serves as a remediation strategy for the non-ideal scenario. The model is explicitly trained in Stage C to develop summary inference capabilities, aiming to improve robustness when evaluated under Stage D, a non-ideal setting.
\subsection{Constructing Noisy Evaluation Sets}
\label{sec:construct_noisy}
In real-world applications, noise-free environments are rare. To better assess the robustness of RL-fine-tuned models, we need to evaluate their reasoning abilities under noisy scenarios.

We categorize input noise into two types: \textbf{fine-grained noise} and \textbf{irrelevant contextual information} (see Figure~\ref{fig:problem_statement}). To support quantitative evaluation, we construct two noisy test sets—\textbf{FineTest} and \textbf{FilterTest}—derived from the original test sets of each dataset, as summarized in Table~\ref{tab:rq2-stats}.

\textbf{FineTest} is built from the original \textbf{TestA} set for each dataset: \textit{Math12k}, \textit{MathReasoning}, \textit{Mathverse}, and \textit{MathVision}. Following the structure in Problem 2 of Figure~\ref{fig:problem_statement}, we inject synthetic distractors (e.g., $x_a^1$, $y$) into the question text. The construction adheres to two principles: (1) only fine-grained distractors are inserted, and (2) the original meaning of the question must be preserved.

\textbf{FilterTest} is derived from each dataset’s \textbf{TestB} set. As illustrated in Problem 3 of Figure~\ref{fig:problem_statement}, we prepend irrelevant context—such as unrelated news or weather updates—to the question. For instance, a math problem about purchasing beer may begin with text about OpenAI or weather conditions. Again, construction follows two principles: (1) the original semantic intent of the question must remain unchanged, and (2) only irrelevant contextual interruptions are added.

Both \textbf{FineTest} and \textbf{FilterTest} are constructed via carefully designed crowdsourcing activities, with strict quality control to ensure consistency and reliability across datasets. Further details on the crowdsourcing procedure can be found in Appendix Section Noisy Evaluation Sets.
\subsection{Robustness to Noise}
Inspired by results in brain science~\citep{tune2021neural,daume2024control}, robustness to noise is a key indicator of human reasoning capacity. We evaluate whether RL-fine-tuned models can similarly maintain reasoning performance under noisy conditions.

As shown in the right part of Figure~\ref{fig:model_figure}, the training process is divided into Stage~E and Stage~G depending on whether the input prompt includes a guiding example for reasoning in the presence of noise. Stage~G includes such an example, while Stage~E does not. Similarly, the evaluation process is divided into Stage~F and Stage~H: Stage~H includes a guiding example, while Stage~F does not. Furthermore, Stage~F is subdivided into Stage~F(a) and Stage~F(b), where Stage~F(b) involves noisy input questions, and Stage~F(a) contains noise-free questions.

Based on different combinations of training and evaluation stages, we define the following RL-fine-tuned model and its variants:

\textbf{Model-StageEF(a)}: The standard RL-fine-tuned model evaluated under noise-free conditions.

\textbf{Model-StageEF(b)}: The same RL-fine-tuned model evaluated under noisy conditions, without any additional guidance. This setup is used to assess the model's robustness to noise.

To address the performance degradation observed in noisy settings, we propose two remediation strategies that incorporate a guiding example during training and/or evaluation. Details of the guiding example can be found in Appendix Section Robustness to Noise.

The guiding example is designed to:
\begin{enumerate}
    \item Highlight relevant information,
    \item Identify and isolate noisy or irrelevant content,
    \item Demonstrate clear reasoning processes.
\end{enumerate}

Given that large models are pre-trained on extensive datasets, we assume they possess basic reasoning capabilities. This approach seeks to activate these capabilities and integrate them into a more robust and advanced reasoning process.

\textbf{Model-StageEH}: This variant introduces a guiding example during evaluation to help the RL-fine-tuned model reason effectively in noisy scenarios by leveraging multiple basic reasoning skills.

\textbf{Model-StageGH}: This variant adds the guiding example to both training and evaluation prompts, encouraging the model to learn robust reasoning patterns throughout its learning process.

\section{Experiments}
\subsubsection{Research Questions}
To assess the reasoning performance of LMs fine-tuned with RL in non-ideal scenarios, we investigate the following research questions (\textbf{RQ}s): \begin{enumerate}
    \item \textbf{Summary Inference:} Can LMs fine-tuned with RL reason over multiple possibilities and perform summary inference when presented with diverse information?

    \item \textbf{Fine-grained Noise Suppression:} Can LMs fine-tuned with RL ignore fine-grained noise and focus on relevant information to reason effectively and reach correct conclusions?

    \item \textbf{Contextual Filtering:} Can LMs fine-tuned with RL disregard irrelevant contextual information and reason effectively to reach valid conclusions?
\end{enumerate}

\subsubsection{Baselines}
We conducted extensive experiments with the following state-of-the-art (SOTA) models:
\begin{itemize}
  \item \textbf{Large Vision-Language Model (LVLM):}
  \begin{itemize}
    \item Qwen 2.5-VL-7B-Instruct~\cite{bai2025qwen2}
  \end{itemize}
  \item \textbf{Large Language Models (LLMs):}
  \begin{itemize}
    \item Llama 3.1-8B-Instruct~\cite{costarelli2024meta}
    \item Qwen 3-14B~\cite{yang2025qwen3}
    \item Mistral-Small-24B-Instruct-2501~\cite{caminha2025evaluating}
  \end{itemize}
\end{itemize}

All baselines are instruction-tuned versions released by the open-source community.
Additionally, we evaluated variants of these models to further analyze their reasoning capabilities.
\subsection{Datasets}
To ensure an objective evaluation, we include only data samples with a single, explicitly defined ground-truth answer.

To address RQ 1, we use \textbf{multiple choice questions} that involve multiple possibilities and require a conclusive answer, aligning with the characteristics of RQ 1.

The data samples used to evaluate RQ1 for LLMs are drawn from the following sources:
\begin{itemize}
    \item \textbf{CommonsenseQA}~\citep{talmor2019commonsenseqa}: a multiple choice question dataset requiring various types of commonsense knowledge to identify the correct answers.
    \item \textbf{Ceval-exam}~\citep{huang2023c}: a multiple choice dataset covering 52 diverse academic disciplines.
\end{itemize}

For the LVLM, the data sources include:
\begin{itemize}
    \item  \textbf{MathVision}~\citep{wang2024measuring}: a dataset in the mathematics domain, where each sample contains an image, a textual description, and a question.
    \item \textbf{WeThink-Multimodal-Reasoning-120K}~\citep{yang2025wethink} (\textbf{WeThink}): a general-purpose multimodal reasoning dataset covering multiple domains.
    \item \textbf{SciVQA}\footnote{\url{https://huggingface.co/datasets/katebor/SciVQA}}: a dataset containing questions derived from English scientific publications.
\end{itemize}
\begin{table}[t]
\centering
\resizebox{0.65\columnwidth}{!}{%
\begin{tabular}{cccc}
\hline
\textbf{Dataset} & \textbf{\#Train} & \textbf{\#Validation} & \textbf{\#Test} \\ \hline
CommonsenseQA    & 2000             & 500                   & 1000            \\
Ceval-exam       & 700              & 246                   & 400             \\
MathVision       & 700              & 200                   & 500             \\
WeThink          & 3000             & 1000                  & 2000            \\
SciVQA           & 2000             & 400                   & 1000            \\ \hline
\end{tabular}%
}
\caption{Statistics of data samples used for Research Question 1. \# denotes the number of data samples.}
\label{tab:rq1-stats}
\end{table}

The statistics for the data samples used in RQ1 are presented in Table~\ref{tab:rq1-stats}.

To analyze RQ 2 and 3, we use \textbf{both} multiple choice and open-ended questions to better reflect the real-world question formats. To maintain objectivity, we focus on math questions as they provide fixed answers in both content and format.

The data samples used to evaluate RQ 2 and 3 for LLMs are drawn from the following sources:
\begin{itemize}
    \item \textbf{Math12k}\footnote{\url{https://huggingface.co/datasets/hiyouga/math12k}} and \textbf{MathReasoning}\footnote{\url{https://huggingface.co/datasets/notbadai/math_reasoning}}: Datasets in the mathematics domain, where each sample has a textual question.
\end{itemize}

For the LVLM, the data sources include:
\begin{itemize}
    \item \textbf{Mathverse}~\citep{zhang2024mathverse} and \textbf{MathVision}: Datasets in the mathematics domain, where each sample contains an image, a textual description, and a question.
\end{itemize}
\begin{table}[t]
\centering
\resizebox{\columnwidth}{!}{%
\begin{tabular}{ccccccc}
\hline
\textbf{Dataset} & \textbf{\#Train} & \textbf{\#Val} & \textbf{\#TestA} & \textbf{\#TestB} & \textbf{\#FineTest} & \textbf{\#FilterTest} \\ \hline
Math12k       & 2500 & 500 & 388  & 745  & 388  & 745  \\
MathReasoning & 3000 & 500 & 468 & 1000 & 468 & 1000 \\
Mathverse     & 280  & 90  & 115  & 114  & 115  & 114  \\
MathVision    & 1400 & 400 & 336  & 480  & 336  & 480  \\ \hline
\end{tabular}%
}
\caption{Statistics of data samples used for RQ 2 and RQ 3. \# denotes the number of samples; Val refers to the validation set. For RQ 2, TestA is the original test set and FineTest is the corresponding noisy test set. For RQ 3, TestB is the original test set and FilterTest is the corresponding noisy test set.}
\label{tab:rq2-stats}
\end{table}

We rely on crowdsourcing to construct two noisy evaluation sets: FineTest for RQ 2 and FilterTest for RQ 3. Each derived from the original test set corresponding to its research question. Detailed methodology is provided in Section Constructing Noisy Evaluation Sets. The statistics for the data samples used in RQ 2 and 3 are presented in Table~\ref{tab:rq2-stats}.

\subsection{Evaluation Metrics}
Each data sample has a single, explicitly defined ground-truth answer. To ensure reproducibility, we use greedy decoding during evaluation. We adopt Pass@1 accuracy, denoted as \textbf{ACC}, as our evaluation metric. Pass@1 accuracy~\citep{shinn2023reflexion} measures the proportion of instances in which the model's single generated response matches the correct answer.
\subsection{Implementation Details}
We organize the implementation details into general and model-specific settings.

\textbf{General Settings.} We use DeepSpeed 0.17.0, TRL 0.18.2, Transformers 4.52.4, and Python 3.12. All experiments are conducted on 4 NVIDIA A100 80GB GPUs. The maximum prompt length is set to 2000 tokens and the maximum output length to 700 tokens. During training, we use temperature = 0.9 and top-$p$ = 1.0. Evaluation is performed using greedy decoding. We randomly fix seeds to 42, 41, and 40 for reproducibility. The model is evaluated on the validation set every 30 steps, with early stopping triggered after 4 evaluations without improvement. We employ the Adam optimizer with a warm-up ratio of 0.05. For GRPO training, 6 generations are sampled per input.

Additional general and model-specific implementation details are provided in Appendix Section Implementation Details.
\subsection{Summary Inference Evaluation}
\begin{table}[t]
\centering
\resizebox{0.7\columnwidth}{!}{%
\begin{tabular}{ccc}
\hline
\textbf{Model}    & \textbf{Commonsense} & \textbf{Ceval-exam} \\ \hline
\textbf{Llama3.1} & 67.30                & 33.75               \\
Llama3.1-StageAB       & 69.80±0.50           & 49.42±0.66          \\
Llama3.1-StageAD       & 64.87±0.37           & 39.00±0.54          \\
Llama3.1-StageCD       & 69.70±0.08           & 49.41±0.72          \\ \hline
\textbf{Mistral}  & 76.60                & 65.50               \\
Mistral-StageAB        & 78.53±0.25           & 67.08±0.42          \\
Mistral-StageAD        & 74.10±0.08           & 63.17±0.24          \\
Mistral-StageCD        & 76.47±0.09           & 65.08±0.82          \\ \hline
\textbf{Qwen3}    & 67.40                 & 71.50                \\
Qwen3-StageAB          & 74.80±0.42           & 74.00±0.36          \\
Qwen3-StageAD          & 75.40±0.16           & 70.00±0.46          \\
Qwen3-StageCD          & 77.5±0.52            & 69.12±0.28          \\ \hline
\end{tabular}%
}
\caption{Evaluation accuracy (\%) of RL-fine-tuned LLMs and their variants on the summary inference scenario. Results are averaged over three runs. 
Results of Models and their stage variants on the same task.  
The Model and StageAB omit summary inference style.  
StageAD applies summary inference only at evaluation, whereas StageCD uses it in both training and evaluation. Table~\ref{tab:rq1-lvlm} uses the same \textbf{discrimination}. 
}
\label{tab:rq1-llms}
\end{table}
 
\begin{table}[t]
\centering
\resizebox{0.8\columnwidth}{!}{%
\begin{tabular}{cccc}
\hline
\textbf{Model}      & \textbf{MathVision} & \textbf{SciVQA} & \textbf{WeThink} \\ \hline
\textbf{Qwen2.5-VL} & 11.00               & 50.80            & 64.10            \\
Qwen2.5-VL-StageAB  & 15.73±0.57          & 60.17±0.60      & 64.93±0.25       \\
Qwen2.5-VL-StageAD  & 10.07±1.39          & 48.80±0.50      & 57.57±0.45       \\
Qwen2.5-VL-StageCD  & 14.53±1.27          & 53.17±1.02      & 58.22±0.87       \\ \hline
\end{tabular}%
}
\caption{Evaluation accuracy (\%) of RL-fine-tuned LVLM and its variants on the summary inference scenario. Results are averaged over three runs.}
\label{tab:rq1-lvlm}
\end{table}

Summary inference is an important reasoning task that merits systematic empirical evaluation. We analyze it across five datasets using three LLMs and one LVLM, yielding the following key findings: (\textbf{1}) Under ideal evaluation settings, which reflect the prevailing practice in current research, RL fine-tuning substantially improves reasoning performance. As shown in Table~\ref{tab:rq1-llms} and Table~\ref{tab:rq1-lvlm}, StageAB variants consistently outperform their base models, highlighting RL's effectiveness in reinforcing correct reasoning paths during training. 
 (\textbf{2}) In contrast, under non-ideal scenarios requiring explicit summarization of all possibilities, RL-fine-tuned models show marked performance degradation. The consistent drop from StageAB to StageAD across all models reveals limited advanced reasoning ability. Additional information confuses the models rather than enhancing performance. (\textbf{3}) To address this limitation, we introduce a remediation strategy in StageCD. The model is trained to consider multiple possibilities and rewarded only for correct, well-reasoned answers. This restores performance under non-ideal conditions and enhances summarization and advanced reasoning. However, the observed performance gap also highlights that RL-fine-tuned models still face significant deficits in advanced reasoning. (\textbf{4}) Larger models consistently outperform smaller ones. As shown in Table~\ref{tab:rq1-llms}, Mistral (24B) and Qwen3 (14B) surpass LLaMA3.1 (8B),
 suggesting that a larger number of parameters enhances reasoning by increasing knowledge capacity and representational power.

\subsection{Fine-grained Noise Suppression Evaluation}
\begin{table}[t]
\centering
\resizebox{0.78\columnwidth}{!}{%
\begin{tabular}{ccc}
\hline
\textbf{Model}        & \multicolumn{2}{c}{\textbf{Fine-grained Noise Suppression}} \\ \hline
                      & \textbf{Mathverse}           & \textbf{MathVision}          \\ \hline
\textbf{Qwen2.5-VL}   & 19.13                        & 22.10                        \\
Qwen2.5-VL-StageEF(a) & 44.64±1.79                   & 22.76±0.14                   \\
Qwen2.5-VL-StageEF(b) & 36.52±1.88                   & 20.53±0.42                   \\
Qwen2.5-VL-StageEH    & 36.81±0.82                   & 19.44±0.28                   \\
Qwen2.5-VL-StageGH    & 43.77±2.05                   & 21.23±0.74                   \\ \hline
\textbf{Model}                 & \multicolumn{2}{c}{\textbf{Contextual Filtering}}           \\ \hline
                      & \textbf{Mathverse}           & \textbf{MathVision}          \\ \hline
\textbf{Qwen2.5-VL}   & 22.61                        & 13.40                        \\
Qwen2.5-VL-StageEF(a) & 43.26±0.38                   & 20.00±0.75                   \\
Qwen2.5-VL-StageEF(b) & 42.98±0.71                   & 19.86±0.35                   \\
Qwen2.5-VL-StageEH    & 39.76±1.09                   & 20.24±0.79                   \\
Qwen2.5-VL-StageGH    & 44.15±1.80                   & 16.26±0.86                   \\ \hline
\end{tabular}%
}
\caption{Evaluation accuracy (\%) of RL-fine-tuned LVLM and its variants on two robustness to noise scenarios. Results are averaged over three runs.
Obtained results of the Model and its stage variants on the same task.  
Model and StageEF(a) are evaluated on the noise-free test set, whereas StageEF(b), StageEH, and StageGH are evaluated on the noisy test set.  
StageGH has a guiding example during both training and evaluation, while StageEH receives it only during evaluation.  
Tables~\ref{tab:llms-fine-grained} and~\ref{tab:llms-contextual-filtering} use the same \textbf{discrimination}. 
}
\label{tab:qwen25-r23}
\end{table}
Fine-grained noise suppression is essential for enabling advanced reasoning without being misled by subtle irrelevant information. We conduct an empirical analysis across four datasets using one LVLM and three LLMs. The results, presented in Table~\ref{tab:qwen25-r23} and Table~\ref{tab:llms-fine-grained}, reveal the following findings: 
(\textbf{1}) RL fine-tuning generally enhances reasoning abilities on the ideal scenario TestA (Table~\ref{tab:rq2-stats}), as shown by comparing performances between Model and Model-StageEF(a) pairs. However, Llama3.1 exhibits performance degradation, indicating difficulty in learning from sparse rewards. In this case, the advantage term $A_i$ in Equation~\ref{eq:grpo_objective} approaches zero, causing the KL divergence $\mathrm{D}_{\mathrm{KL}}(\pi_\theta \| \pi_{\text{ref}})$ to update the policy in unhelpful directions, thereby impairing its original reasoning ability. 
(\textbf{2}) Model-StageEF(b) is evaluated on the non-ideal scenario FineTest, which introduces fine-grained noise to TestA (Table~\ref{tab:rq2-stats}). Comparing multiple Model-StageEF(b) and Model-StageEF(a) pairs reveals a clear performance drop, particularly for Mistral. This indicates that RL fine-tuned models fail to effectively suppress fine-grained noise, reflecting limited advanced reasoning capabilities. 

(\textbf{3}) Model-StageEH and Model-StageGH are two remediation strategies designed for the fine-grained noise scenario FineTest. As shown in Table~\ref{tab:qwen25-r23} and Table~\ref{tab:llms-fine-grained}, both methods outperform Model-StageEF(b) in at least one case, confirming our assumption: introducing examples during training and/or evaluation can activate and integrate basic reasoning capabilities into more advanced reasoning processes. 

Specifically, Qwen3-StageEH outperforms Qwen3-StageGH, suggesting that Qwen3 possesses strong basic reasoning skills that can be effectively activated during evaluation. In contrast, training with examples in Qwen3-StageGH updates parameters in ways that may impair the model’s originally latent reasoning abilities. For Llama3.1, the trend is reversed. Llama3.1-StageEH performs poorly, indicating its basic reasoning capabilities are insufficient for the fine-grained noise scenario. However, Llama3.1-StageGH, which incorporates examples during training, improves performance by encouraging the learning of fine-grained distinctions and updating parameters accordingly. Mistral-StageEH and Mistral-StageGH perform similarly, indicating limited sensitivity to the training stage intervention.

For the vision-language model Qwen2.5-VL, StageEH provides little improvement, as recognizing fine-grained noise requires image-text comparison that cannot be effectively trained during evaluation alone. In contrast, Qwen2.5-VL-StageGH introduces examples during training, enabling the model to learn to distinguish fine-grained noise and reinforcing this ability through parameter updates.

\begin{table}[t]
\centering
\resizebox{0.75\columnwidth}{!}{%
\begin{tabular}{ccc}
\hline
\textbf{Model}      & \multicolumn{2}{c}{\textbf{Fine-grained Noise Suppression}} \\ \hline
                    & \textbf{Math12k}          & \textbf{MathReasoning}          \\ \hline
\textbf{Mistral}    & 63.47                     & 69.20                           \\
Mistral-StageEF(a)  & 69.58±0.92                & 74.47±0.41                      \\
Mistral-StageEF(b)  & 60.70±0.13                & 72.22±1.14                      \\
Mistral-StageEH     & 61.21±0.13                & 72.34±0.17                      \\
Mistral-StageGH     & 60.57±0.79                & 74.57±0.94                      \\ \hline
\textbf{Qwen3}      & 51.73                     & 65.12                           \\
Qwen3-StageEF(a)    & 56.58±0.31                & 67.33±0.90                      \\
Qwen3-StageEF(b)    & 50.69±0.32                & 66.17±0.71                      \\
Qwen3-StageEH       & 58.33±0.32                & 67.25±0.84                      \\
Qwen3-StageGH       & 56.70±0.21                & 66.52±0.66                      \\ \hline
\textbf{Llama3.1}   & 50.40                     & 46.70                            \\
Llama3.1-StageEF(a) & 41.13±0.38                & 35.80±0.75                      \\
Llama3.1-StageEF(b) & 39.43±0.58                & 35.41±0.89                      \\
Llama3.1-StageEH    & 40.72±0.77                & 32.26±0.76                      \\
Llama3.1-StageGH    & 44.33±0.77                & 36.32±0.57                      \\ \hline
\end{tabular}%
}
\caption{Evaluation accuracy (\%) of RL-fine-tuned LLMs and their variants on the fine-grained noise suppression scenario. Results are averaged over three runs.}
\label{tab:llms-fine-grained}
\end{table}
\subsection{Contextual Filtering Evaluation}
\begin{table}[t]
\centering
\resizebox{0.72\columnwidth}{!}{%
\begin{tabular}{ccc}
\hline
\textbf{Model}      & \multicolumn{2}{c}{\textbf{Contextual Filtering}} \\ \hline
                    & \textbf{Math12k}     & \textbf{MathReasoning}     \\ \hline
\textbf{Mistral}    & 63.87                & 69.30                      \\
Mistral-StageEF(a)  & 69.02±1.04           & 74.83±0.59                 \\
Mistral-StageEF(b)  & 66.31±0.73           & 66.10±0.96                 \\
Mistral-StageEH     & 65.77±0.85           & 66.3±0.79                  \\
Mistral-StageGH     & 65.99±0.72           & 70.33±0.85                 \\ \hline
\textbf{Qwen3}      & 53.20                & 66.69                      \\
Qwen3-StageEF(a)    & 56.53±0.68           & 68.26±0.39                 \\
Qwen3-StageEF(b)    & 57.27±0.44           & 68.93±0.38                 \\
Qwen3-StageEH       & 63.80±0.46           & 70.83±0.17                 \\
Qwen3-StageGH       & 62.97±0.09           & 69.67±0.29                 \\ \hline
\textbf{Llama3.1}   & 50.27                & 47.50                       \\
Llama3.1-StageEF(a) & 40.80±0.22           & 42.90±0.57                 \\
Llama3.1-StageEF(b) & 42.89±0.75           & 37.96±0.86                 \\
Llama3.1-StageEH    & 46.89±1.89           & 37.60±0.79                 \\
Llama3.1-StageGH    & 49.60±0.74           & 39.30±0.83                 \\ \hline
\end{tabular}%
}
\caption{Evaluation accuracy (\%) of RL-fine-tuned LLMs and their variants on the contextual filtering scenario. Results are averaged over three runs.}
\label{tab:llms-contextual-filtering}
\end{table}
Contextual filtering is crucial for removing irrelevant information and focusing on key contents. We conducted an empirical study on four datasets. The results, reported in Table~\ref{tab:qwen25-r23} and Table~\ref{tab:llms-contextual-filtering}, yield the following observations:
(\textbf{1}) On the ideal scenario TestB (Table~\ref{tab:rq2-stats}), a comparison between the Models and Model-StageEF(a) variants shows that RL fine-tuning improves reasoning in every model except Llama3.1. Llama3.1 fails to learn from sparse rewards, leading to policy collapse and a decline in accuracy.
(\textbf{2}) Model-StageEF(b) is evaluated on the non-ideal scenario FilterTest, which augments TestB (Table~\ref{tab:rq2-stats}) with irrelevant context. Relative to Model-StageEF(a), most models maintain their performance, demonstrating effective contextual filtering during reasoning, with the sole exception of Mistral. The robustness of Qwen3, Qwen2.5-VL, and Llama3.1 is likely due to pre-training on diverse corpora that include text summarisation tasks, fostering an ability to discard irrelevant information.

(\textbf{3}) Model-StageEH and Model-StageGH address contextual filtering. Tables~\ref{tab:qwen25-r23} and~\ref{tab:llms-contextual-filtering} show that both strategies improve, or at least preserve, accuracy on FilterTest relative to Model-StageEF(b), confirming that examples can activate and integrate latent contextual filtering skills. Model-StageEH supplies examples only at evaluation time, prompting the model to discard contextual noise without modifying its parameters. In contrast, Model-StageGH injects examples during training, guiding optimisation to focus on key content. For the vision-language model Qwen2.5-VL, gains appear only with StageGH, indicating that parameter updates are required to align textual and visual features. A domain shift issue on the MathVision subset is also observed, which is common in deep learning models.
\section{Conclusion}
In this work, we discover that while RL can enhance the reasoning abilities of LLMs, these models still struggle in non-ideal scenarios
where humans excel. 
To investigate the underlying reasons, we propose a new research direction and introduce three realistic, non-ideal scenarios. 
Using a representative RL fine-tuning method, GRPO, we evaluate the performance of three LLMs and a LVLM across these tasks. 
Experimental results show that while our proposed scenario-specific remediation methods improve reasoning, existing RL approaches still exhibit critical limitations in non-ideal settings. 
These findings provide a foundation for future research.

\appendix

\bibliography{aaai2026}

\end{document}